\pdfoutput=1
\documentclass{article}

\usepackage[preprint]{neurips_2026}

\usepackage[utf8]{inputenc} % allow utf-8 input
\usepackage[T1]{fontenc}    % use 8-bit T1 fonts
\usepackage{url}            % simple URL typesetting
\usepackage{booktabs}       % professional-quality tables
\usepackage{amsmath}        % math environments and extensible arrows
\usepackage{amsfonts}       % blackboard math symbols
\usepackage{nicefrac}       % compact symbols for 1/2, etc.
\usepackage{microtype}      % microtypography
\usepackage{xcolor}         % colors
\usepackage{graphicx}       % figures
\usepackage{float}          % fixed figure placement for case studies
\usepackage{placeins}       % keep floats close to their sections

\usepackage{booktabs}
\usepackage{multirow}
\usepackage{makecell}
\usepackage{graphicx}
\usepackage{tikz}
\usepackage{pgfplots}
\usepgfplotslibrary{groupplots}
\pgfplotsset{compat=1.18}

\usepackage{listings}
\usepackage{hyperref}       % hyperlinks

\definecolor{keywordcolor}{rgb}{0.7, 0.1, 0.1}
\definecolor{tacticcolor}{rgb}{0.0, 0.1, 0.6}
\definecolor{commentcolor}{rgb}{0.4, 0.4, 0.4}
\definecolor{symbolcolor}{rgb}{0.0, 0.1, 0.6}
\definecolor{sortcolor}{rgb}{0.1, 0.5, 0.1}
\definecolor{attributecolor}{rgb}{0.7, 0.1, 0.1}

\lstdefinelanguage{lean}{
mathescape=false,
texcl=false,
morekeywords=[1]{
import, prelude, protected, private, noncomputable, definition, meta, renaming,
hiding, parameter, parameters, begin, constant, constants,
lemma, variable, variables, theory,
print, theorem, example,
open, as, export, override, axiom, axioms, inductive, with,
structure, record, universe, universes,
alias, help, precedence, reserve, declare_trace, add_key_equivalence,
match, infix, infixl, infixr, notation, postfix, prefix, instance,
eval, reduce, check, end, this,
using, using_well_founded, namespace, section,
attribute, local, set_option, extends, include, omit, class,
raw, replacing,
calc, have, show, suffices, by, in, at, let, forall, Pi, fun,
exists, if, dif, then, else, assume, obtain, from, classical, register_simp_ext, unless, break, continue,
mutual, do, def, run_cmd, const,
partial, mut, where, macro, syntax, deriving,
return, try, catch, for, macro_rules, declare_syntax_cat, abbrev},
morekeywords=[2]{Sort, Type, Prop},
morekeywords=[3]{
assumption,
apply, intro, intros, allGoals,
generalize, clear, revert, done, exact, exact_mod_cast, mod_cast,
refine, repeat, cases, rcases, obtain, rewrite, rw, rwa,
simp, simpa, simp_all, dsimp, contradiction,
constructor, injection, induction,
by_cases, by_contra, contrapose, specialize, use,
rfl, subst, split, left, right,
linarith, nlinarith, omega, positivity, norm_num,
norm_cast, push_cast, push_neg,
ring, ring_nf, field_simp, ext, funext,
},
literate=
{α}{{\ensuremath{\mathrm{\alpha}}}}1
{β}{{\ensuremath{\mathrm{\beta}}}}1
{γ}{{\ensuremath{\mathrm{\gamma}}}}1
{δ}{{\ensuremath{\mathrm{\delta}}}}1
{ε}{{\ensuremath{\mathrm{\varepsilon}}}}1
{ζ}{{\ensuremath{\mathrm{\zeta}}}}1
{η}{{\ensuremath{\mathrm{\eta}}}}1
{θ}{{\ensuremath{\mathrm{\theta}}}}1
{ι}{{\ensuremath{\mathrm{\iota}}}}1
{κ}{{\ensuremath{\mathrm{\kappa}}}}1
{μ}{{\ensuremath{\mathrm{\mu}}}}1
{ν}{{\ensuremath{\mathrm{\nu}}}}1
{ξ}{{\ensuremath{\mathrm{\xi}}}}1
{π}{{\ensuremath{\mathrm{\mathnormal{\pi}}}}}1
{ρ}{{\ensuremath{\mathrm{\rho}}}}1
{σ}{{\ensuremath{\mathrm{\sigma}}}}1
{τ}{{\ensuremath{\mathrm{\tau}}}}1
{φ}{{\ensuremath{\mathrm{\varphi}}}}1
{χ}{{\ensuremath{\mathrm{\chi}}}}1
{ψ}{{\ensuremath{\mathrm{\psi}}}}1
{ω}{{\ensuremath{\mathrm{\omega}}}}1
{Γ}{{\ensuremath{\mathrm{\Gamma}}}}1
{Δ}{{\ensuremath{\mathrm{\Delta}}}}1
{Θ}{{\ensuremath{\mathrm{\Theta}}}}1
{Λ}{{\ensuremath{\mathrm{\Lambda}}}}1
{Σ}{{\ensuremath{\mathrm{\Sigma}}}}1
{Φ}{{\ensuremath{\mathrm{\Phi}}}}1
{Ξ}{{\ensuremath{\mathrm{\Xi}}}}1
{Ψ}{{\ensuremath{\mathrm{\Psi}}}}1
{Ω}{{\ensuremath{\mathrm{\Omega}}}}1
{ℵ}{{\ensuremath{\aleph}}}1
{≤}{{\ensuremath{\leq}}}1
{≥}{{\ensuremath{\geq}}}1
{≠}{{\ensuremath{\neq}}}1
{≈}{{\ensuremath{\approx}}}1
{≡}{{\ensuremath{\equiv}}}1
{≃}{{\ensuremath{\simeq}}}1
{≤}{{\ensuremath{\leq}}}1
{≥}{{\ensuremath{\geq}}}1
{∂}{{\ensuremath{\partial}}}1
{∆}{{\ensuremath{\triangle}}}1 % or \laplace?
{∫}{{\ensuremath{\int}}}1
{∑}{{\ensuremath{\mathrm{\Sigma}}}}1
{Π}{{\ensuremath{\mathrm{\Pi}}}}1
{⊥}{{\ensuremath{\perp}}}1
{∞}{{\ensuremath{\infty}}}1
{∂}{{\ensuremath{\partial}}}1
{∓}{{\ensuremath{\mp}}}1
{±}{{\ensuremath{\pm}}}1
{×}{{\ensuremath{\times}}}1
{⊕}{{\ensuremath{\oplus}}}1
{⊗}{{\ensuremath{\otimes}}}1
{⊞}{{\ensuremath{\boxplus}}}1
{∇}{{\ensuremath{\nabla}}}1
{√}{{\ensuremath{\sqrt}}}1
{⬝}{{\ensuremath{\cdot}}}1
{•}{{\ensuremath{\cdot}}}1
{∘}{{\ensuremath{\circ}}}1
{⁻}{{\ensuremath{^{-}}}}1
{▸}{{\ensuremath{\blacktriangleright}}}1
{∧}{{\ensuremath{\wedge}}}1
{∨}{{\ensuremath{\vee}}}1
{¬}{{\ensuremath{\neg}}}1
{⊢}{{\ensuremath{\vdash}}}1
{⟨}{{\ensuremath{\langle}}}1
{⟩}{{\ensuremath{\rangle}}}1
{↦}{{\ensuremath{\mapsto}}}1
{←}{{\ensuremath{\leftarrow}}}1
{<-}{{\ensuremath{\leftarrow}}}1
{→}{{\ensuremath{\rightarrow}}}1
{↔}{{\ensuremath{\leftrightarrow}}}1
{⇒}{{\ensuremath{\Rightarrow}}}1
{⟹}{{\ensuremath{\Longrightarrow}}}1
{⇐}{{\ensuremath{\Leftarrow}}}1
{⟸}{{\ensuremath{\Longleftarrow}}}1
{∩}{{\ensuremath{\cap}}}1
{∪}{{\ensuremath{\cup}}}1
{⋃}{{\ensuremath{\bigcup}}}1
{⊂}{{\ensuremath{\subseteq}}}1
{⊆}{{\ensuremath{\subseteq}}}1
{⊄}{{\ensuremath{\nsubseteq}}}1
{⊈}{{\ensuremath{\nsubseteq}}}1
{⊃}{{\ensuremath{\supseteq}}}1
{⊇}{{\ensuremath{\supseteq}}}1
{⊅}{{\ensuremath{\nsupseteq}}}1
{⊉}{{\ensuremath{\nsupseteq}}}1
{∈}{{\ensuremath{\in}}}1
{∉}{{\ensuremath{\notin}}}1
{∋}{{\ensuremath{\ni}}}1
{∌}{{\ensuremath{\notni}}}1
{∅}{{\ensuremath{\emptyset}}}1
{∖}{{\ensuremath{\setminus}}}1
{†}{{\ensuremath{\dag}}}1
{ℕ}{{\ensuremath{\mathbb{N}}}}1
{ℤ}{{\ensuremath{\mathbb{Z}}}}1
{ℝ}{{\ensuremath{\mathbb{R}}}}1
{ℚ}{{\ensuremath{\mathbb{Q}}}}1
{ℂ}{{\ensuremath{\mathbb{C}}}}1
{⌞}{{\ensuremath{\llcorner}}}1
{⌟}{{\ensuremath{\lrcorner}}}1
{⦃}{{\ensuremath{\{\!|}}}1
{⦄}{{\ensuremath{|\!\}}}}1
{‖}{{\ensuremath{\|}}}1
{₁}{{\ensuremath{_1}}}1
{₂}{{\ensuremath{_2}}}1
{₃}{{\ensuremath{_3}}}1
{₄}{{\ensuremath{_4}}}1
{₅}{{\ensuremath{_5}}}1
{₆}{{\ensuremath{_6}}}1
{₇}{{\ensuremath{_7}}}1
{₈}{{\ensuremath{_8}}}1
{₉}{{\ensuremath{_9}}}1
{₀}{{\ensuremath{_0}}}1
{ᵢ}{{\ensuremath{_i}}}1
{ⱼ}{{\ensuremath{_j}}}1
{ₐ}{{\ensuremath{_a}}}1
{¹}{{\ensuremath{^1}}}1
{²}{{\ensuremath{^2}}}1
{ᶜ}{{\ensuremath{^c}}}1
{ₙ}{{\ensuremath{_n}}}1
{ₘ}{{\ensuremath{_m}}}1
{ₚ}{{\ensuremath{_p}}}1
{↑}{{\ensuremath{\uparrow}}}1
{↓}{{\ensuremath{\downarrow}}}1
{...}{{\ensuremath{\ldots}}}1
{·}{{\ensuremath{\cdot}}}1
{▸}{{\ensuremath{\triangleright}}}1
{Σ}{{\color{symbolcolor}\ensuremath{\Sigma}}}1
{Π}{{\color{symbolcolor}\ensuremath{\Pi}}}1
{∀}{{\color{symbolcolor}\ensuremath{\forall}}}1
{∃}{{\color{symbolcolor}\ensuremath{\exists}}}1
{λ}{{\color{symbolcolor}\ensuremath{\mathrm{\lambda}}}}1
{\$}{{\color{symbolcolor}\$}}1
{:=}{{\color{symbolcolor}:=}}1
{=}{{\color{symbolcolor}=}}1
{<|>}{{\color{symbolcolor}<|>}}1
{<\$>}{{\color{symbolcolor}<\$>}}1
{+}{{\color{symbolcolor}+}}1
{*}{{\color{symbolcolor}*}}1,
morecomment=[s][\color{commentcolor}]{/-}{-/},
morecomment=[l][\itshape \color{commentcolor}]{--},
showstringspaces=false,
keepspaces=true,
morestring=[b]",
morestring=[d],
tabsize=3,
extendedchars=false,
sensitive=true,
breaklines=true,
breakatwhitespace=true,
basicstyle=\ttfamily\small,
captionpos=b,
columns=[l]fullflexible,
identifierstyle={\ttfamily\color{black}},
keywordstyle=[1]{\ttfamily\color{keywordcolor}},
keywordstyle=[2]{\ttfamily\color{sortcolor}},
keywordstyle=[3]{\ttfamily\color{tacticcolor}},
keywordstyle=[4]{\ttfamily\color{attributecolor}},
stringstyle=\ttfamily,
commentstyle={\ttfamily\footnotesize },
}

\lstdefinestyle{leanlisting}{
  language=lean,
  basicstyle=\ttfamily\scriptsize,
  breaklines=true,
  breakatwhitespace=false,
  columns=fullflexible,
  keepspaces=true,
  showstringspaces=false,
  tabsize=2,
  frame=single,
  framerule=0.2pt,
  xleftmargin=0.5em,
  xrightmargin=0.5em,
  captionpos=b,
  inputencoding=utf8,
  extendedchars=true,
  alsoletter={_},
  keywordstyle=[1]{\ttfamily\color{keywordcolor}},
  keywordstyle=[2]{\ttfamily\color{sortcolor}},
  keywordstyle=[3]{\ttfamily\color{tacticcolor}},
  keywordstyle=[4]{\ttfamily\color{attributecolor}},
  commentstyle={\ttfamily\itshape\color{commentcolor}},
  stringstyle={\ttfamily\color{black}},
  identifierstyle={\ttfamily\color{black}}
}

\lstdefinestyle{promptlisting}{
  basicstyle=\ttfamily\scriptsize,
  breaklines=true,
  breakatwhitespace=false,
  columns=fullflexible,
  keepspaces=true,
  showstringspaces=false,
  tabsize=2,
  frame=single,
  framerule=0.2pt,
  xleftmargin=0.5em,
  xrightmargin=0.5em,
  captionpos=b,
  inputencoding=utf8,
  extendedchars=true,
  literate=
    {ℕ}{{$\mathbb{N}$}}1 {ℚ}{{$\mathbb{Q}$}}1 {ℝ}{{$\mathbb{R}$}}1
    {ℂ}{{$\mathbb{C}$}}1 {ℤ}{{$\mathbb{Z}$}}1
    {α}{{$\alpha$}}1 {β}{{$\beta$}}1 {γ}{{$\gamma$}}1 {ε}{{$\varepsilon$}}1
    {ι}{{$\iota$}}1 {λ}{{$\lambda$}}1 {μ}{{$\mu$}}1 {ξ}{{$\xi$}}1
    {π}{{$\pi$}}1 {ω}{{$\omega$}}1
    {∀}{{$\forall$}}1 {∃}{{$\exists$}}1 {¬}{{$\neg$}}1 {∧}{{$\wedge$}}1
    {∨}{{$\vee$}}1 {∈}{{$\in$}}1 {∉}{{$\notin$}}1 {∩}{{$\cap$}}1
    {∪}{{$\cup$}}1 {⋃}{{$\bigcup$}}1 {∑}{{$\sum$}}1 {∏}{{$\prod$}}1
    {∫}{{$\int$}}1 {≤}{{$\le$}}1 {≥}{{$\ge$}}1 {≠}{{$\ne$}}1
    {⊂}{{$\subset$}}1 {⊆}{{$\subseteq$}}1 {⊢}{{$\vdash$}}1 {⊤}{{$\top$}}1
    {→}{{$\to$}}1 {←}{{$\leftarrow$}}1 {↔}{{$\leftrightarrow$}}1
    {↦}{{$\mapsto$}}1 {↑}{{$\uparrow$}}1 {∘}{{$\circ$}}1 {×}{{$\times$}}1
    {∂}{{$\partial$}}1 {∣}{{$\mid$}}1 {⁻}{{$^{-}$}}1
    {⟨}{{$\langle$}}1 {⟩}{{$\rangle$}}1 {⟪}{{$\langle\!\langle$}}1
    {⟫}{{$\rangle\!\rangle$}}1 {⦃}{{$\lbrace$}}1 {⦄}{{$\rbrace$}}1
    {‖}{{$\Vert$}}1 {·}{{$\cdot$}}1 {▸}{{$\triangleright$}}1
    {¹}{{$^{1}$}}1 {²}{{$^{2}$}}1 {₁}{{$_1$}}1 {₂}{{$_2$}}1
    {₃}{{$_3$}}1 {₄}{{$_4$}}1 {ₙ}{{$_n$}}1 {ᶜ}{{$^c$}}1
    {ᶠ}{{$^f$}}1 {ᵢ}{{$_i$}}1 {ⱼ}{{$_j$}}1 {ˢ}{{$^s$}}1
    {–}{{--}}1 {—}{{--}}1 {…}{{...}}1 {§}{{\S}}1
    {─}{{-}}1 {│}{{|}}1 {┌}{{+}}1 {┐}{{+}}1 {└}{{+}}1 {┘}{{+}}1
    {├}{{+}}1 {┤}{{+}}1 {┬}{{+}}1 {┴}{{+}}1 {┼}{{+}}1
    {✓}{{\checkmark}}1 {✅}{{[ok]}}1 {✗}{{x}}1 {❌}{{[x]}}1
    {⚠}{{!}}1 {️}{{}}1 {�}{{?}}1
}

\title{Proof-Refactor: Refactoring Generated Formal Proofs into Modular Artifacts}

\author{%
  Yiming Fu \\
  Department of Mathematics \\
  Southern University of Science and Technology \\
  \And
  Peixuan Liu \\
  Department of Mathematics \\
  Southern University of Science and Technology \\
  \AND
  Zichen Wang \\
  School of Mathematical Sciences \\
  Peking University\\
  \And
  Kun Yuan\thanks{\begin{tabular}[t]{@{}l@{}}Corresponding author. Email address: \texttt{kunyuan@pku.edu.cn}.\\
  Code: \href{https://github.com/pelicanhere/proof-refactor}{https://github.com/pelicanhere/proof-refactor}.\end{tabular}} \\
  Center for Machine Learning Research \\ 
  Peking University \\
}

\begin{document}
\raggedbottom

\maketitle

\begin{abstract}
While Large Language Models (LLMs) have shown strong performance in generating formal proofs, their outputs often remain less readable, modular, maintainable, and reusable than proofs in mature formal mathematics libraries. We argue that this gap stems in part from the compile-first objective implicit in most proof-generation pipelines, which encourages monolithic or ad hoc proof scripts rather than library-quality artifacts.
Existing approaches to proof-quality improvement often rely on explicit, computable optimization objectives. In practice, however, the most tractable and experimentally validated objectives are largely length-based, while higher-level qualities such as readability, modularity, maintainability, and reusability are difficult to reduce to reliable automatic metrics. Instead of optimizing proof improvement against a single proxy metric, we take a process-guided approach inspired by human proof-refactoring workflows. We propose an agentic framework \textbf{Proof-Refactor} that decomposes proof refactoring into four phases: extracting candidate proof fragments, designing helper declarations, formally proving the extracted and designed components, and repairing the original proof using the verified components. On generated Lean proofs from PutnamBench and Putnam2025, Proof-Refactor improves rubric-based refactoring scores over a strong Claude Code refactoring baseline, with the largest gains in signature quality and human readability. These results suggest that process-guided refactoring can improve proof structure without treating proof length as the primary objective.
\end{abstract}

\section{Introduction}

Formal theorem proving with proof assistants such as Lean~\citep{lean} and
Isabelle~\citep{isabelle} produces machine-verifiable mathematical proofs. This
is useful for verified mathematics and for training LLMs, since proof checking
provides a precise reward signal for reinforcement learning (RL). Recent systems
have made progress on university-level mathematics, including PutnamBench
\citep{tsoukalas2024putnambench} and Putnam2025 problems
\citep{axiommath2025seeingwhy}.

However, proof-assistant verification also induces a narrow objective: a
generated proof is rewarded primarily for compiling. This binary feedback
ensures correctness, but it does not by itself encourage readability,
modularity, maintainability, or library integration. An agent may therefore
produce a passing proof that leaves reusable arguments inlined, introduces
problem-specific definitions instead of using existing ones, or relies on heavy
automation and enlarged resource limits rather than factoring the argument into
intermediate declarations. This reflects a mismatch between what is easy to
check and the broader qualities expected of library-quality formal mathematics.

Existing approaches to improving proof quality
\citep{ahuja2024improver,gu2025proofoptimizertraininglanguagemodels} often rely
on computable objectives. In practice, the most reliable automatic objectives
are largely length-based, while qualities such as readability, modularity,
maintainability, and reusability are difficult to measure. Length-based
objectives can also favor direct compression (golfing) over semantically
meaningful helper declarations.

This distinction mirrors human formalization practice. Humans often first
obtain a compiling proof and then refactor it. The central step is not only to
reduce length, but to identify reusable proof fragments, generalize them into
declarations that fit the surrounding library, prove these declarations from the
existing fragments, and rewrite the original proof to call them. Proof golfing
is a legitimate part of refactoring in this broad sense, but it is not the whole
process: focusing only on golfing misses the helper-oriented workflow that turns
local arguments into reusable library components.

We therefore view proof refactoring as a structured process rather than
end-to-end optimization against a proxy metric. Proof-Refactor does not include
a dedicated length-minimization stage; instead, it focuses on extracting,
generalizing, and reusing intermediate proof components. We propose
\textbf{Proof-Refactor}, a Claude Code-based agentic framework that decomposes
Lean proof refactoring into four phases: extracting candidate proof fragments,
designing helper declarations, proving the extracted and designed components,
and repairing the original proof with the verified components. The framework
integrates \texttt{lean-lsp-mcp}~\citep{lean-lsp-mcp}, a custom Lean 4
\texttt{extract} tactic for turning proof fragments into standalone theorems,
and external assistance for selecting fragments and designing helper
declarations. Experiments on generated PutnamBench and Putnam2025 proofs provide
evidence that Proof-Refactor improves refactoring quality over the
baseline, with the clearest gains in helper-signature quality, proof structure,
reuse of mathematical infrastructure, and human readability.

\section{Related work}

\paragraph{Proof Optimization.}
A related line of work improves formal proofs through computable objectives,
especially proof length
\citep{ahuja2024improver,gu2025proofoptimizertraininglanguagemodels}.
\citet{ahuja2024improver} combine LLM search with retrieval-augmented
generation (RAG) and Chain-of-States, while
ProofOptimizer~\citep{gu2025proofoptimizertraininglanguagemodels} trains models
with length-filtered data and length-based RL rewards. These systems show that
post-generation proof improvement is feasible and can produce synthetic
training data for models. Our work targets a complementary form of
improvement: helper declarations are the central object of a structured
refactoring pipeline, rather than a byproduct of optimizing a metric.

\paragraph{Proof Extraction.}
Another related direction is extracting artifacts from existing formal proofs.
REFACTOR \citep{zhou2024refactor} trains a graph neural network (GNN) to identify
proof-tree nodes that form extractable theorem components in Metamath
\citep{megill2019metamath}, then standardizes the predicted components into
valid theorems.
\citet{xin2025automateddiscoverytacticlibraries} study a more symbolic approach,
discovering reusable tactics by searching for recurring, collapsible subgraphs
in Tactic Dependence Graphs. These works are closely related to proof
refactoring, but extraction alone tends to preserve the form and specificity of
the original fragments rather than lifting them into more general, library-style
artifacts. This motivates the dedicated helper-design phase in Proof-Refactor,
which explicitly revises extracted scaffolds into cleaner and more reusable
declarations.

\paragraph{Agentic Systems for Theorem Proving.}
Hilbert \citep{varambally2026hilbertrecursivelybuildingformal} and Numina-Lean-Agent
\citep{liu2026numinaleanagentopengeneralagentic} are open-source agentic systems
for formal theorem proving. Hilbert combines informal reasoning, theorem
retrieval, formal proving, and verifier feedback to recursively decompose hard
goals into subgoals. Numina-Lean-Agent builds on Claude Code and
Numina-Lean-MCP, using tools for Lean interaction, retrieval, informal proving,
and external-model consultation through a discussion partner. These systems are
related to the draft-sketch-prove paradigm~\citep{jiang2023draftsketchproveguiding},
where informal reasoning supplies high-level proof sketches that guide formal
provers toward a complete proof.

We adopt two system-level ideas from this line: using Claude Code with
\texttt{lean-lsp-mcp} as the Lean interaction substrate, and allowing the agent
to consult an external model. However, our use of external assistance is not an
informal-to-formal sketching pipeline. Proof-Refactor starts from an existing
Lean proof, and external assistance serves as an auxiliary reasoning channel for
refactoring decisions over the current formal context, rather than only producing
informal proof sketches. Thus, while it is related in spirit to
Numina-Lean-Agent's discussion partner, the goal is to support proof refactoring
rather than to produce a proof plan for theorem proving.

\section{Refactoring: Task and Evaluation}

\paragraph{Task Definition.}
Let \(E\) denote the ambient formal environment available to a
declaration, including imported library declarations. Given
a verified declaration \((S,p)\), where \(S\) is its formal signature and \(p\)
is its original proof, the initial data are \((E,p)\), satisfying
\(E \vdash p : S\).

We formulate an individual refactoring as a constrained transformation
\((E,p) \leadsto_{\mathrm{refactor}} (E',p')\):
\begin{equation}
(E,p) \leadsto_{\mathrm{refactor}} (E',p') \quad \text{s.t.} \quad
\begin{cases}
  E' \vdash p' : S \\
  (E',p') \succ_{\mathcal{Q}} (E,p)
\end{cases}
\end{equation}

Here, \(E \vdash p : S\) means: under the
environment \(E\), the proof \(p\) of the formal signature \(S\) complies.
The relation \((E',p') \succ_{\mathcal{Q}} (E,p)\) states that the refactored
proof is preferred to the original proof according to the quality preference
\(\mathcal{Q}\). The same preference can also be applied to a finite set of
candidate refactorings to compare or rank them against the original proof.
In this paper, we only consider transformations where \(E' = E \cup H\), where
\(H\) is a set of newly introduced declarations; the original environment
\(E\) is left unchanged.

\paragraph{Evaluation.} We instantiate \(\mathcal{Q}\) with a rubric-based
quality score that compares candidate refactorings against the original proof. The judge protocol and
length-based diagnostics are described in Section~\ref{sec:experiments}.

\FloatBarrier
\section{Proof-Refactor: Process-Driven Refactoring}

\begin{figure}[H]
  \centering
  \includegraphics[width=\linewidth]{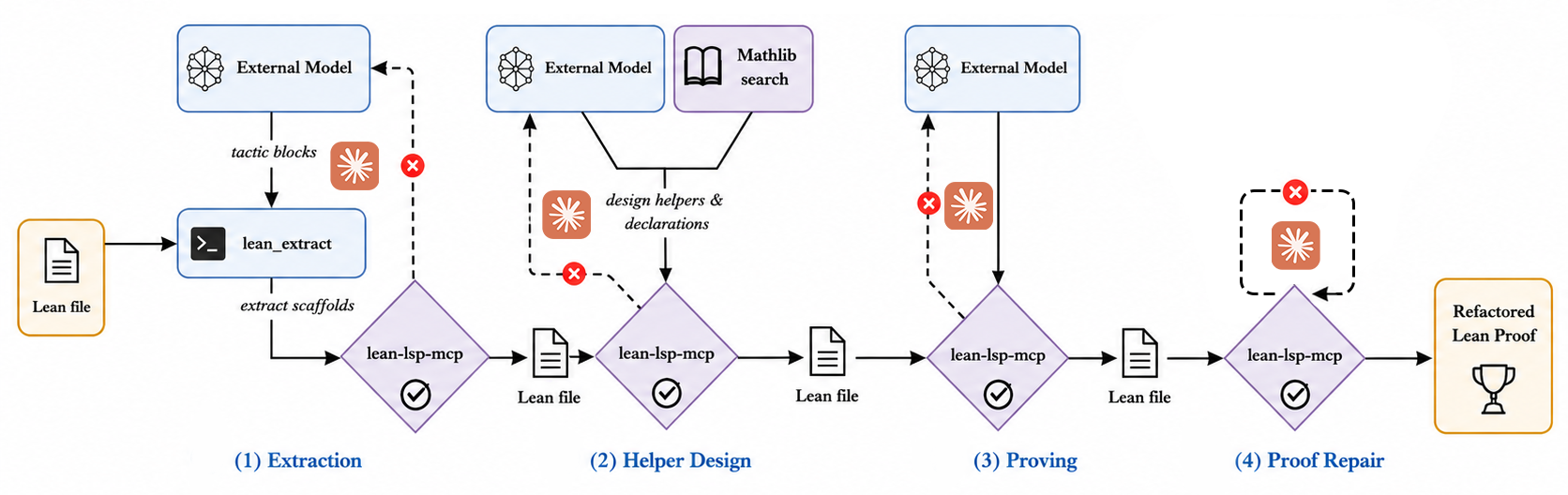}
  \caption{Overview of the Proof-Refactor pipeline.}
  \label{fig:proof-refactor-pipeline}
\end{figure}

As shown in Figure~\ref{fig:proof-refactor-pipeline}, Proof-Refactor is a four-phase agentic
framework for Lean proof refactoring, built on Claude Code and \texttt{lean-lsp-mcp}. The pipeline
can call external models through an independent-context interface. We extend
\texttt{lean-lsp-mcp} with \texttt{lean\_extract}, a custom extraction tool based
on the Lean tactic \texttt{extract}, which can extract proof fragments into
standalone theorems.

\subsection{Lean-LSP-MCP}
We use \texttt{lean-lsp-mcp}, which is the Model Context Protocol (MCP) server
for Lean. We implement a prompt about the \texttt{lean-lsp-mcp} API modified
from \citet{lean4-skills}. Concretely, we use these tools in three main ways:
\begin{itemize}
  \item \textbf{Context inspection.} We use
  \texttt{lean\_file\_outline}, \texttt{lean\_hover\_info}, \texttt{lean\_goal},
  and \texttt{lean\_diagnostic\_messages} to inspect file structure,
  declaration information, proof states, and Lean error messages.
  \item \textbf{Declaration search.} The system uses
  \texttt{lean\_local\_search} to retrieve local and Mathlib declarations by
  name, \texttt{lean\_leansearch} for natural-language search
  over Mathlib, and \texttt{lean\_loogle} for type-pattern search.

  \item \textbf{Proof interaction.} We use
  \texttt{lean\_multi\_attempt} to test multiple candidate tactics in parallel,
  \texttt{lean\_code\_actions} to resolve Lean suggestions, and
  \texttt{lean\_run\_code} to execute standalone Lean snippets. Each attempted
  change is checked using \texttt{lean\_goal} and
  \texttt{lean\_diagnostic\_messages}.
\end{itemize}

\FloatBarrier
\subsection{The Extract Tactic}
The \texttt{extract} tactic is a custom Lean tactic that we implement for proof
extraction. It extracts a tactic block inside a proof into a standalone theorem.
We design the tactic for two main cases: (1) the tactic block either closes the
goal or leaves a goal definitionally equal to the original goal, in which case
the extracted theorem keeps the original goal as its goal; and (2) the
tactic block leaves one remaining goal that is not definitionally equal to the
original goal, in which case the extracted theorem states the implication from
the post-block goal to the pre-block goal.
The extracted theorem keeps only the variables actually used by
the tactic block and removes unused variables.
This design supports both \texttt{have}-style subproofs and contiguous tactic
blocks that transform the goal to another goal. It differs from the
Mathlib tactic \texttt{extract\_goal}, which extracts a theorem from the current
goal at the line where the tactic is inserted, but does not support case (2) and
cannot clean up some variables properly. We
integrate the \texttt{extract} tactic into \texttt{lean-lsp-mcp} through
\texttt{lean\_extract}, a script that automatically inserts \texttt{extract}
wrappers into the proof and inserts the extracted theorem into the file. We call
this inserted declaration a \emph{scaffold}: an intermediate top-level
theorem whose signature is produced by \texttt{extract}. A scaffold is not yet a
final helper declaration; it is a structured artifact passed to the later
phases.

Scaffolds play three roles in the pipeline. First, they provide concrete
material for helper design by exposing the local assumptions, goals, and proof
fragments extracted from the owner declaration, i.e., the original declaration
that contains the tactic block. Second, they help the proving phase localize the
task to smaller lemmas rather than reasoning about the whole owner proof at
once. Third, during proof repair, the inserted \texttt{extract} wrapper records
the original tactic block, and the scaffold proof body provides a verified
guide for reorganizing the owner proof. This makes repair a guided
reconstruction step rather than an in-place edit inside the original
declaration.

\begin{figure}[!b]
  \centering
  \begin{minipage}[t]{0.40\linewidth}
    \centering
    {\small Before extraction}\\[0.25em]
    \includegraphics[width=\linewidth]{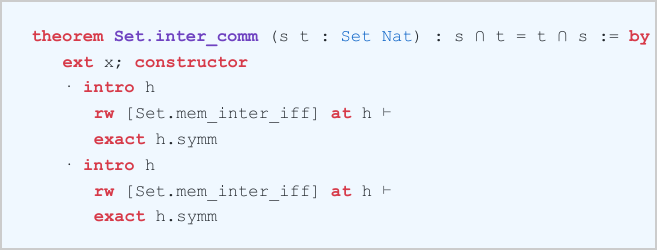}
  \end{minipage}
  \hfill
  \begin{minipage}[t]{0.56\linewidth}
    \centering
    {\small After extraction}\\[0.25em]
    \includegraphics[width=\linewidth]{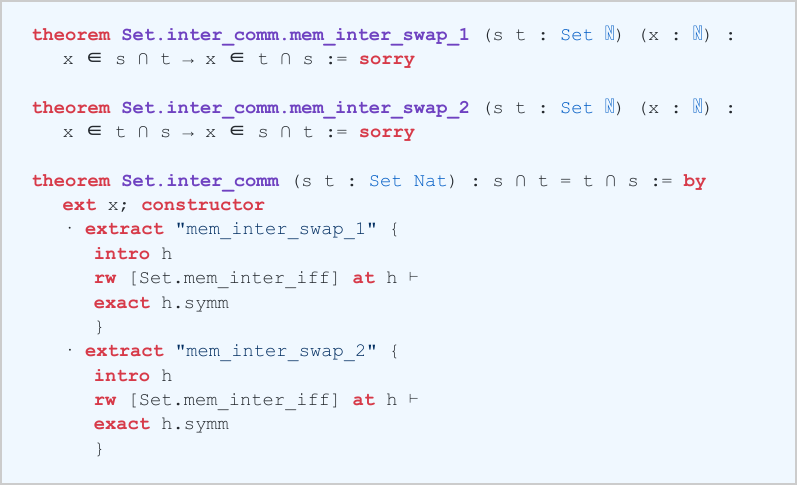}
  \end{minipage}
  \caption[Example proof before and after extraction.]{Example proof before and
  after extraction. The two inserted declarations are scaffolds, and
  \texttt{Set.inter\_comm} is the owner declaration.}
  \label{fig:extract-before-after}
\end{figure}

\subsection{Four-phase Pipeline}
Proof-Refactor decomposes refactoring into four sequential phases. Each phase is
a separate Claude Code session with \texttt{lean-lsp-mcp} that does not inherit the previous
session's context; the Lean file produced by the previous phase is the only
artifact passed forward.

\paragraph{External Assistance.}
We implement a CLI that allows the pipeline to call external models with an independent context.
The purpose is to decouple high-level refactoring reasoning from low-level interaction with Lean:
the external model can reason about which fragments are mathematically meaningful and how they should
be generalized, while the Claude Code agent remains responsible for editing, checking, and repairing
Lean code. In practice, external assistance is used primarily in the extraction and helper design
phases, where it helps select proof fragments and propose reusable helper declarations. The pipeline
can also consult it as a secondary fallback when revising helper signatures or proving unusually
difficult obligations.

\paragraph{Phase 1: Extraction.}
The extraction phase begins by asking external assistance to read the original
Lean proof and return extraction suggestions, each corresponding to a tactic
block. The agent then calls \texttt{lean\_extract} to turn the suggestions into scaffolds. It checks each scaffold with
\texttt{lean-lsp-mcp}; if any errors, the agent repairs them before
moving to the next phase. The inserted \texttt{extract} wrapper is left in the
file for later phases.

\paragraph{Phase 2: Helper Design.}
The design phase reads each scaffold together with its owner declaration, and
then groups scaffolds by similarity. The agent writes the scaffolds, owner
declarations, relevant Mathlib declarations, and search evidence into a Markdown
file, then asks the external model to design helpers for each group. The agent reviews
the proposed helper signatures against design principles. If needed, the agent may make
a secondary call to external assistance for revision. The resulting helper declarations are
inserted into a Lean file, and each declaration is annotated with dependency
information to organize the proof. In this dependency structure, scaffolds are
temporary: no other declaration is allowed to depend on them. The agent then
proceeds to the next phase.

\paragraph{Phase 3: Proving.}
The proving phase fills in each \texttt{sorry} in the Lean file. The agent reads the
scaffolds, helper declarations, owner declarations, and extraction sites, and then proves each theorem
guided by the dependency annotations. It may consult the external model when it encounters a
difficult proof obligation, but this is rarely used in practice: most goals are small lemmas, and
the original proof fragment usually provides the proof idea, even when its code cannot be reused
verbatim. Once all scaffolds and helper declarations have been proved, the agent proceeds to the
next phase.

\paragraph{Phase 4: Proof Repair.}
Finally, the agent repairs the original proof by following the dependency
annotations and invoking the helper declarations. At each extraction site, it uses
the corresponding scaffold body as guidance for rewriting, while ensuring that the
repaired proof does not call the scaffold itself. The agent then removes the
\texttt{extract} wrapper and the scaffolds, and adds mathematical annotations.

\section{Experiments}
\label{sec:experiments}
We evaluate Proof-Refactor on generated Lean proofs from PutnamBench and Putnam2025. Each evaluation
instance is a self-contained Lean file whose only external dependency is mathlib. Our evaluation
focuses on refactoring quality rather than direct length-based optimization; accordingly, we do not
emphasize proof golfing as an explicit objective.

\subsection{Experimental setup}

\paragraph{Datasets.} We evaluate on generated Lean proofs from PutnamBench
\citep{tsoukalas2024putnambench} and Putnam2025. For PutnamBench, we use Claude Code with
DeepSeek-V4-Pro \citep{deepseekai2026deepseekv4} to generate candidate proofs and select 96
proofs whose lengths mostly range from 50 to 500 lines, with a median of approximately 160 lines.
We additionally evaluate on 12 Putnam2025 proofs generated by \citet{axiommath2025seeingwhy}.

\paragraph{Models and tools.} Except for the evaluation judge described below, all refactoring agents
are built on Claude Code and \texttt{lean-lsp-mcp}. Both our method and the baseline use Claude Opus
4.6~\citep{ClaudeOpus46} as the underlying model.
The baseline is Claude Code with the \texttt{lean4:refactor} command from
lean4-skills~\citep{lean4-skills}. When our method invokes an external model, we use Gemini 3.1
Pro~\citep{deepmind2026gemini31pro}. We consider this a strong refactoring baseline: the
lean4-skills agent can inspect Lean proof states through \texttt{lean\_goal} and use them to guide
proof-state-guided lemma extraction, and its prompts require the extracted lemma signatures to be
clean and general. Moreover, its authors intentionally separate refactoring from proof golfing,
which matches our experimental setup.

\paragraph{Evaluation.} All evaluated files are required to pass Lean verification, so our evaluation
does not score correctness. Instead, we evaluate the quality of the refactored
proof artifact. This is difficult to capture with a single automatic metric:
proof length may reward golfing rather than readability, and the number of
introduced lemmas may reward fake modularity. We therefore use a rubric-based
evaluation protocol that scores refactoring quality along five dimensions:
structure, signature quality, tactic quality, reuse, and human readability.
The rubric explicitly penalizes wrappers, dead helpers, oversized helpers, and
broad tactics that hide the main proof idea, while rewarding meaningful helper
lemmas with natural, general, and locally checkable statements. 

Our primary evaluator is an LLM-as-a-judge agent. To reduce dependence on the
system prompt or tool behavior of the refactoring agents, the judge is
implemented separately from Claude Code. The judge is given the original proof
and the refactored outputs, together with the fixed rubric. It can query
LeanSearch~\citep{gao2025semanticsearchenginemathlib4} to retrieve relevant
Mathlib declarations, allowing it to assess whether a refactoring follows
library style by reusing existing abstractions rather than reinventing them.
For each method, the judge assigns scores to the rubric submetrics, and the
overall score is the average of the non-overall rubric scores.

As a sanity check for the automatic evaluation, we additionally conduct a human
review on 28 valid problems using the same rubric. We compare the judge with the
human review at the theorem level by asking whether they select the same better
refactoring between the two methods. The judge agrees with the human review on
this pairwise ranking for 75.0\% of the sampled theorems.

Finally, following prior work~\citep{gu2025proofoptimizertraininglanguagemodels},
we also report length-based diagnostics. In Table~\ref{tab:score_summary}, we
instantiate length as the word count of the whole Lean file: for an original
file \(F\) and a refactored file \(F'\), we define the word reduction rate as
\(\frac{\mathrm{words}(F)-\mathrm{words}(F')}{\mathrm{words}(F)}\), where
\(\mathrm{words}(\cdot)\) denotes whole-file word count. We also report
the average number of words reduced, \(\mathrm{words}(F) -
\mathrm{words}(F')\), averaged over instances. These length-based diagnostics
are used only as secondary measures, not as the primary measure of refactoring
quality. We freeze all prompts, scripts, model versions, and evaluation settings
before running the experiments.

\subsection{Main results}

Table~\ref{tab:score_summary} reports the main results. Proof-Refactor improves the average quality
score over the baseline by +0.31 on Putnam2025 and +0.45 on PutnamBench. The largest gains appear in
Signature Quality and Human Readability, suggesting that explicitly designing reusable helper
declarations produces proofs that are more natural and better aligned with library conventions.
Structure and Reuse also improve consistently, indicating that the refactored proofs are better
organized and make more effective use of existing mathematical infrastructure.

The length-based diagnostics in Table~\ref{tab:score_summary}, measured by
whole-file word counts, show a more nuanced pattern. Proof-Refactor achieves a
higher word reduction rate on PutnamBench (+2.5pp over the baseline), but a
lower reduction rate on Putnam2025 (-3.8pp). This is consistent with our
objective: the pipeline is not optimized directly for proof golfing, and its
quality gains primarily come from reorganizing proofs around reusable helper
declarations. The gain in Tactic Quality is therefore smaller but remains
positive, reflecting our focus on high-level proof architecture rather than
tactic-level compression.

\begin{table}[H]
\centering
\caption{Comparison between baseline and ours.}
\label{tab:score_summary}
\resizebox{0.94\textwidth}{!}{%
\begin{tabular}{ccccc}
\toprule
\textbf{Dataset} & \textbf{Metric} & \textbf{Baseline} & \textbf{Proof-Refactor} & \textbf{Improvement} \\
\midrule
\multirow[c]{8}{*}{\makecell[c]{Putnam2025\\(12 problems)}}
& \textbf{Average score} & 3.96 & \textbf{4.27} & \textbf{+0.31} \\
& Structure & 4.01 & 4.31 & +0.29 \\
& \textbf{Signature Quality} & 3.76 & \textbf{4.28} & \textbf{+0.51} \\
& Tactic Quality & 4.04 & 4.22 & +0.18 \\
& Reuse & 4.00 & 4.29 & +0.29 \\
& Human Readability & 3.99 & 4.26 & +0.28 \\
& Word reduction rate & 8.3\% & 4.5\% & -3.8pp \\
& Avg.\ words reduced & 625.9 & 337.2 & -288.7 \\
\midrule
\multirow[c]{8}{*}{\makecell[c]{PutnamBench\\(96 problems)}}
& \textbf{Average score} & 3.75 & \textbf{4.19} & \textbf{+0.45} \\
& Structure & 3.80 & 4.21 & +0.41 \\
& \textbf{Signature Quality} & 3.46 & \textbf{4.14} & \textbf{+0.68} \\
& Tactic Quality & 3.97 & 4.19 & +0.22 \\
& Reuse & 3.84 & 4.24 & +0.41 \\
& \textbf{Human Readability} & 3.67 & \textbf{4.18} & \textbf{+0.50} \\
& Word reduction rate & 4.0\% & 6.5\% & +2.5pp \\
& Avg.\ words reduced & 105.5 & 118.1 & +12.6 \\
\bottomrule
\end{tabular}%
}
\end{table}

\subsubsection{Case study: Putnam 1968 B6.}

To better understand the aggregate gains in Table~\ref{tab:score_summary}, we examine Putnam
1968 B6 as a representative case where the main challenge is not shortening tactics, but exposing
the proof's mathematical structure; the complete Lean artifacts for this case are listed in
Appendix~\ref{app:lean-showcase}. The proof naturally decomposes into four steps: assuming, for
contradiction, that there exists a sequence $K_n$ covering all compact sets; choosing a point $q_n$
close to $0$ but not contained in $K_n$ for each $K_n$; using $q_n \to 0$ to construct the compact
set $S=\{0\}\cup\mathrm{range}(q)$; and finally deriving a diagonal contradiction from $q_n\in S$
and $q_n\notin K_n$.

\begin{figure}[!t]
  \centering
  \begin{minipage}[t]{0.84\linewidth}
    \centering
    \includegraphics[width=\linewidth]{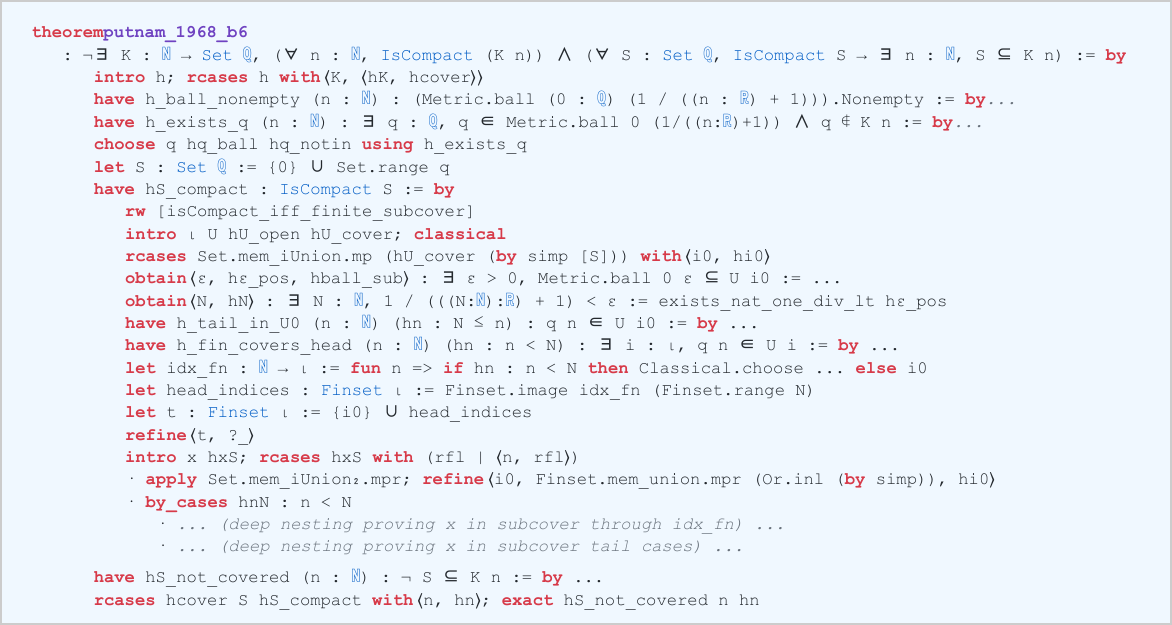}
    {\footnotesize Original generated proof (86 lines)}
  \end{minipage}

  \vspace{0.35em}
  \begin{minipage}[t]{0.84\linewidth}
    \centering
    \includegraphics[width=\linewidth]{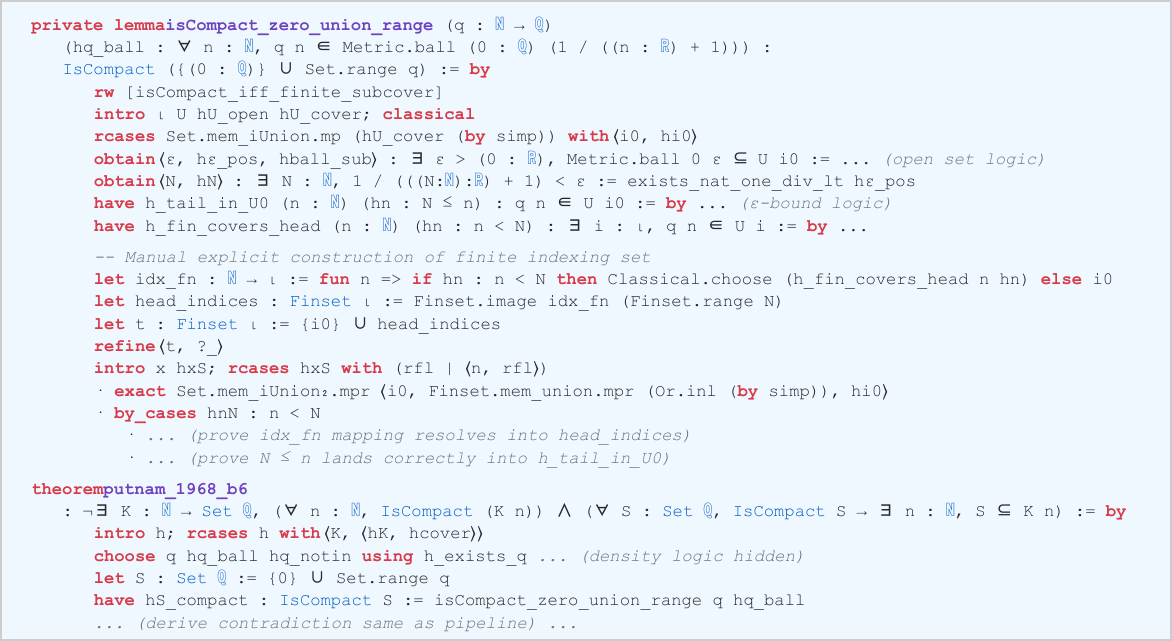}
    {\footnotesize Baseline refactoring (84 lines)}
  \end{minipage}

  \vspace{0.35em}
  \begin{minipage}[t]{0.84\linewidth}
    \centering
    \includegraphics[width=\linewidth]{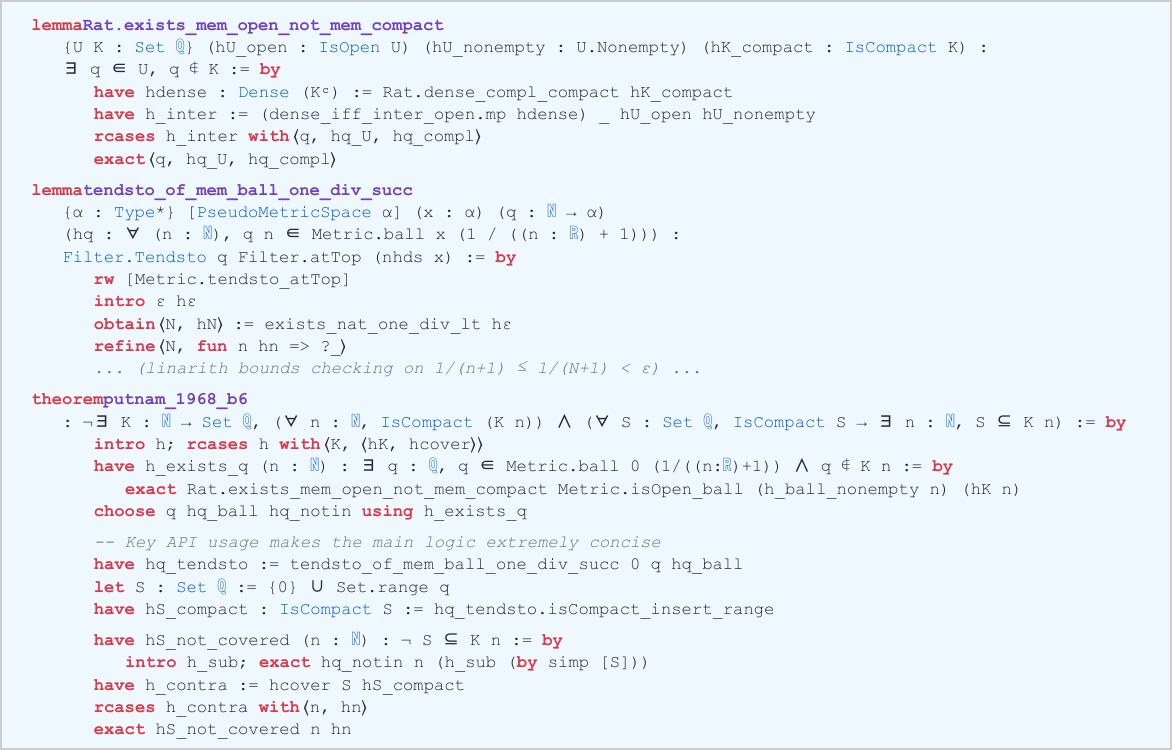}
    {\footnotesize Proof-Refactor (52 lines)}
  \end{minipage}
  \caption{Case study on Putnam 1968 B6. Snippets are lightly abridged and
  reformatted to fit the main text.}
  \label{fig:putnam-1968-b6-case-study}
\end{figure}

While the original proof interleaves these steps with substantial low-level topological details---such as expanding the compactness proof into an explicit open-cover argument---the baseline merely extracts this into a highly specialized helper lemma, \path|isCompact_zero_union_range|. Because this baseline lemma fixes the space to $\mathbb{Q}$, the limit point to \(0\), the radius to \(1/(n+1)\), and the conclusion directly to the compactness of \(\{0\}\cup\mathrm{range}(q)\), it acts as a direct translation of the original proof structure; moreover, its proof body is itself largely copied from the corresponding part of the original proof. This highlights a primary failure mode of the baseline: because it attempts an end-to-end refactor within a single coding session, it is overburdened. Consequently, it struggles to propose clean, library-style lemma signatures or discover simpler proofs, despite prompt-level guidance toward cleaner refactorings.

By contrast, Proof-Refactor mitigates these limitations by using a four-phase pipeline with
external assistance. The two lemmas extracted by our approach correspond to core mathematical steps that substantially generalize the original local fragments. Specifically, \path|Rat.exists_mem_open_not_mem_compact| generalizes ``choosing $q_n\notin K_n$ from a specific ball'' into the statement that every nonempty open subset of the rationals contains a point outside any given compact set; meanwhile, \path|tendsto_of_mem_ball_one_div_succ| abstracts the open-cover estimate into a general convergence criterion stating that in any pseudo-metric space, if the $n$-th term of a sequence lies in the ball centered at $x$ with radius $1/(n+1)$, then the sequence converges to $x$. 

This allows the main proof to obtain the compactness of $S$ directly via
\path|hq_tendsto.isCompact_insert_range|. The case study thus explains the pattern observed in
Table~\ref{tab:score_summary}: Proof-Refactor's improvements come primarily from changing the
interface between the main theorem and its helper declarations, rather than from local tactic
optimization. By turning local proof fragments into reusable mathematical facts and then reusing
appropriate library declarations, the pipeline directly improves Signature Quality, Structure, and
Reuse, while also making the resulting proof easier to inspect for human readability. This also explains
why the gain in Tactic Quality is smaller: Proof-Refactor targets high-level proof architecture, not
tactic-level proof golfing.

\FloatBarrier
\subsection{Ablations and analysis}

We run an ablation that removes external assistance while keeping all other settings fixed. The
ablation is evaluated on Putnam2025 and a 48-problem subset of PutnamBench. All metrics are measured
as improvements over the original generated proofs, and we report the ablated score minus the
full-pipeline score as the ablation gap. This ablation removes the primary external calls used for
fragment selection and helper design, as well as the rare fallback calls used for helper revision or
difficult proof obligations.

\begin{table}[H]
\centering
{\scriptsize
\renewcommand{\arraystretch}{0.88}
\caption{Ablation study on removing external assistance.}
\label{tab:ablation_design}
\begin{tabular}{ccccc}
\toprule
Group & Metric & Full Pipeline & Ablation & Gap \\
\midrule

\multirow[c]{3}{*}{\makecell[c]{Overall\\shared problems}}
& Pipeline improvement & +0.31 & -0.55 & -0.86 \\
& Pipeline win rate & 39.1\% & 30.4\% & -8.7pp \\
& Pipeline loss rate & 52.2\% & 69.6\% & +17.4pp \\

\midrule

\multirow[c]{5}{*}{\makecell[c]{Dimension\\improvement}}
& Structure & +0.41 & -0.93 & -1.34 \\
& Signature Quality & +0.68 & -0.30 & -0.98 \\
& Human Readability & +0.50 & -0.67 & -1.17 \\
& Reuse & +0.41 & -0.50 & -0.91 \\
& Tactic Quality & +0.22 & -0.37 & -0.59 \\

\bottomrule
\end{tabular}
}
\end{table}

Table~\ref{tab:ablation_design} shows that removing external assistance changes the pipeline from a
positive average improvement (+0.31) to a negative one (-0.55), increases the loss rate from 52.2\%
to 69.6\%, and reduces the win rate by 8.7 percentage points. The largest drops are in Structure
(-1.34), Human Readability (-1.17), and Signature Quality (-0.98). This supports the importance of
decoupling high-level reasoning from Lean interaction. Without external assistance, the same agent
context must simultaneously manage tool calls, Lean diagnostics, local proof repair, search results,
and the higher-level task of designing reusable helper declarations. This mixed context can become
overloaded and poorly focused, which directly hurts the design phase: helper signatures become less
clean, less general, and less aligned with library conventions. Reuse also drops substantially
(-0.91), suggesting that the ablated pipeline is less able to turn local proof fragments into
reusable components. Tactic Quality drops less sharply (-0.59), consistent with the fact that local
proof repair can still succeed even when the higher-level refactoring architecture is weaker. The
rare fallback calls used during proving are included in the ablation, but the main effect is the loss
of the reasoning--interaction separation used for fragment selection and helper design.

\section{Conclusion}

We introduced Proof-Refactor, a process-guided framework for refactoring generated Lean proofs into
more modular artifacts. Rather than treating proof improvement as direct optimization of a single
proxy such as length, Proof-Refactor decomposes the process into extraction, helper design, proving,
and proof repair. This makes helper declarations the central object of refactoring: local proof
fragments are exposed, generalized into reusable declarations, proved, and then used to reorganize
the original proof. This staged design avoids overloading a single end-to-end agent with extraction,
design, proving, and repair at once, and allows each phase to focus on a more specific refactoring
subproblem.

A key design principle is to decouple high-level refactoring reasoning from low-level Lean
interaction. External assistance is used mainly to reason about meaningful proof fragments and helper
interfaces, while the Claude Code agent remains responsible for Lean editing, checking, proving, and
repair. The ablation study supports this separation: removing external assistance substantially
degrades structure, signature quality, readability, and reuse, suggesting that mixing helper design
with tool-heavy Lean interaction in a single context weakens the refactoring process.

Experiments on generated PutnamBench and Putnam2025 proofs provide evidence that this process
improves refactoring quality over a strong baseline, with the clearest gains in Signature Quality,
Structure, Reuse, and Human Readability. The case study further shows that Proof-Refactor can recover
the conceptual structure of a proof rather than merely copying local proof-engineering details.

\section{Limitations and broader impacts}

\paragraph{Scope.}
This work studies a restricted refactoring setting: each evaluation instance is a self-contained
Lean file, and the refactored environment only adds new declarations. The pipeline does not edit
existing library declarations, reorganize projects, change imports, or perform multi-file
refactoring, so our experiments do not establish project- or library-level scalability. Future work
should extend Proof-Refactor to reason about existing declarations, naming conventions, import
structure, and changes to the surrounding environment, with the long-term goal of making generated
formal proofs suitable for integration into mature libraries such as mathlib~\citep{mathlib}.

\paragraph{Evaluation.}
Our main evaluation uses a rubric-based LLM-as-a-judge protocol, augmented with retrieval from
mathlib and supplemented by human review on a subset. This provides a practical way to evaluate
high-level proof quality, but it is not a substitute for large-scale expert review. The benchmark is
also limited to generated Putnam-style Lean proofs, so the results may not directly generalize to
other mathematical domains, larger developments, or proofs written by human formalizers.

\paragraph{Cost.}
Proof-Refactor uses four sequential phases and external assistance, making it more expensive than a
single end-to-end refactoring pass.

\paragraph{Failure modes.}
Because the pipeline is sequential, errors in early phases can propagate. In particular, failures or
bugs in \texttt{lean\_extract} can produce poor scaffolds, making helper design, proving, and repair
harder. The design phase may also reject useful problem-specific lemmas when it applies the
generality criterion too strictly.

\paragraph{Broader impacts.}
Beyond Proof-Refactor, the \texttt{extract} tactic exposed through \texttt{lean\_extract} may help
theorem-proving systems isolate local contexts and extracted subgoals, and may support synthesis of
formalization data from existing proofs. More broadly, the staged decomposition may inform
code-refactoring systems, though generated helpers should remain draft artifacts before integration
into shared libraries.

% Do not include acknowledgments in an anonymized submission. For the final
% camera-ready version, uncomment this block and disclose funding and competing
% interests as required by NeurIPS.
% \begin{ack}
% Add acknowledgments and disclosure of funding here.
% \end{ack}

% \section*{References}

{\small

\bibliographystyle{plainnat}
\bibliography{references}
}

%%%%%%%%%%%%%%%%%%%%%%%%%%%%%%%%%%%%%%%%%%%%%%%%%%%%%%%%%%%%

\appendix

\section{Additional Artifacts}
\label{app:repro-artifacts}

This appendix records selected Lean examples and the evaluation rubric. We omit the full prompt
suites from the arXiv appendix to keep the source concise; individual listings are captioned by role
rather than by repository path.

\begin{table}[h]
\centering
\caption{Appendix artifact index.}
\label{tab:appendix-artifact-index}
\small
\begin{tabular}{cp{0.34\linewidth}p{0.48\linewidth}}
\toprule
\textbf{ID} & \textbf{Artifact group} & \textbf{Appendix location} \\
\midrule
A1 & Lean showcase artifacts &
Appendix~\ref{app:lean-showcase}, Listings~\ref{lst:case-original-1968b6}--\ref{lst:case-pipeline-1971a2} \\
A2 & Evaluation rubric &
Appendix~\ref{app:evaluation-rubric}, Listing~\ref{lst:evaluation-rubric} \\
\bottomrule
\end{tabular}
\end{table}

\subsection{Lean Showcase Files}
\label{app:lean-showcase}

The listings below provide representative Lean artifacts, including the Putnam 1968 B6 case study
and other representative examples. Each example includes the original generated proof, the
\texttt{lean4:refactor} baseline, and Proof-Refactor.

\begin{lstlisting}[style=leanlisting,caption={Original generated proof for Putnam 1968 B6.},label={lst:case-original-1968b6}]
import Mathlib

open Finset Polynomial Topology Filter Metric

/--
Prove that no sequence $\{K_n\}_{n=0}^{\infty}$ of compact (closed and bounded) sets of rational numbers has the property that every compact set of rational numbers is contained by at least one $K_n$.
-/
theorem putnam_1968_b6
  : ¬∃ K : ℕ → Set ℚ, (∀ n : ℕ, IsCompact (K n)) ∧ (∀ S : Set ℚ, IsCompact S → ∃ n : ℕ, S ⊆ K n) := by
  intro h
  rcases h with ⟨K, ⟨hK, hcover⟩⟩
  have h_ball_nonempty (n : ℕ) : (Metric.ball (0 : ℚ) (1 / ((n : ℝ) + 1))).Nonempty := by
    rw [Metric.nonempty_ball]
    positivity
  have h_exists_q (n : ℕ) : ∃ q : ℚ, q ∈ Metric.ball (0 : ℚ) (1 / ((n : ℝ) + 1)) ∧ q ∉ K n := by
    have hdense : Dense ((K n)ᶜ) := Rat.dense_compl_compact (hK n)
    have hball_open : IsOpen (Metric.ball (0 : ℚ) (1 / ((n : ℝ) + 1))) := Metric.isOpen_ball
    have h_inter := (dense_iff_inter_open.mp hdense) _ hball_open (h_ball_nonempty n)
    rcases h_inter with ⟨q, hq_mem, hq_not⟩
    exact ⟨q, hq_mem, hq_not⟩
  choose q hq_ball hq_notin using h_exists_q
  let S : Set ℚ := {0} ∪ Set.range q
  have hS_compact : IsCompact S := by
    rw [isCompact_iff_finite_subcover]
    intro ι U hU_open hU_cover
    classical
    have h0_mem_S : (0 : ℚ) ∈ S := by
      simp [S]
    have h0_in_cover : (0 : ℚ) ∈ ⋃ i, U i := hU_cover h0_mem_S
    rcases Set.mem_iUnion.mp h0_in_cover with ⟨i0, hi0⟩
    have hi0_open : IsOpen (U i0) := hU_open i0
    obtain ⟨ε, hε_pos, hball_sub⟩ : ∃ ε > (0 : ℝ), Metric.ball (0 : ℚ) ε ⊆ U i0 := by
      rcases (Metric.isOpen_iff.mp hi0_open) (0 : ℚ) hi0 with ⟨ε, hε_pos, hball_sub⟩
      exact ⟨ε, hε_pos, hball_sub⟩
    obtain ⟨N, hN⟩ : ∃ N : ℕ, 1 / (((N : ℕ) : ℝ) + 1) < ε :=
      exists_nat_one_div_lt hε_pos
    have h_tail_in_U0 (n : ℕ) (hn : N ≤ n) : q n ∈ U i0 := by
      have h_bound : dist (q n) (0 : ℚ) < ε := by
        have hqb := Metric.mem_ball.mp (hq_ball n)
        have h_denom : 1 / (((n : ℕ) : ℝ) + 1) ≤ 1 / (((N : ℕ) : ℝ) + 1) := by
          refine (one_div_le_one_div (by positivity) (by positivity)).mpr ?_
          have h_n_ge_N : (N : ℝ) ≤ (n : ℝ) := by exact mod_cast hn
          linarith
        linarith
      have hmem : q n ∈ Metric.ball (0 : ℚ) ε := by
        rw [Metric.mem_ball]
        exact h_bound
      exact hball_sub hmem
    have h_fin_covers_head (n : ℕ) (hn : n < N) : ∃ i : ι, q n ∈ U i := by
      have hqn_mem_S : q n ∈ S := by
        simp [S]
      have hqn_in_cover : q n ∈ ⋃ i, U i := hU_cover hqn_mem_S
      rcases Set.mem_iUnion.mp hqn_in_cover with ⟨i, hi⟩
      exact ⟨i, hi⟩
    let idx_fn : ℕ → ι := fun n =>
      if hn : n < N then Classical.choose (h_fin_covers_head n hn) else i0
    let head_indices : Finset ι :=
      Finset.image idx_fn (Finset.range N)
    let t : Finset ι := {i0} ∪ head_indices
    refine ⟨t, ?_⟩
    intro x hxS
    rcases hxS with (rfl | ⟨n, rfl⟩)
    · apply Set.mem_iUnion₂.mpr
      refine ⟨i0, Finset.mem_union.mpr (Or.inl (by simp)), hi0⟩
    · by_cases hnN : n < N
      · have hi_mem : idx_fn n ∈ head_indices := by
          dsimp [head_indices]
          apply Finset.mem_image.mpr
          exact ⟨n, Finset.mem_range.mpr hnN, rfl⟩
        have hqn_in_U : q n ∈ U (idx_fn n) := by
          dsimp [idx_fn]
          rw [dif_pos hnN]
          exact Classical.choose_spec (h_fin_covers_head n hnN)
        apply Set.mem_iUnion₂.mpr
        exact ⟨idx_fn n, Finset.mem_union.mpr (Or.inr hi_mem), hqn_in_U⟩
      · have hNle : N ≤ n := by omega
        have hx_in_U0 : q n ∈ U i0 := h_tail_in_U0 n hNle
        apply Set.mem_iUnion₂.mpr
        refine ⟨i0, Finset.mem_union.mpr (Or.inl (by simp)), hx_in_U0⟩
  have hS_not_covered (n : ℕ) : ¬ S ⊆ K n := by
    intro h_sub
    have hqn_in_Kn : q n ∈ K n := h_sub (by simp [S])
    exact hq_notin n hqn_in_Kn
  have h_contra := hcover S hS_compact
  rcases h_contra with ⟨n, hn⟩
  exact hS_not_covered n hn
\end{lstlisting}

\begin{lstlisting}[style=leanlisting,caption={Baseline refactoring for Putnam 1968 B6.},label={lst:case-baseline-1968b6}]
import Mathlib

open Finset Polynomial Topology Filter Metric

-- {0} ∪ range(q) is compact when q(n) → 0 at rate 1/(n+1)
private lemma isCompact_zero_union_range (q : ℕ → ℚ)
    (hq_ball : ∀ n : ℕ, q n ∈ Metric.ball (0 : ℚ) (1 / ((n : ℝ) + 1))) :
    IsCompact ({(0 : ℚ)} ∪ Set.range q) := by
  rw [isCompact_iff_finite_subcover]
  intro ι U hU_open hU_cover
  classical
  have h0_mem : (0 : ℚ) ∈ ({(0 : ℚ)} ∪ Set.range q) := by simp
  have h0_in_cover : (0 : ℚ) ∈ ⋃ i, U i := hU_cover h0_mem
  rcases Set.mem_iUnion.mp h0_in_cover with ⟨i0, hi0⟩
  have hi0_open : IsOpen (U i0) := hU_open i0
  obtain ⟨ε, hε_pos, hball_sub⟩ : ∃ ε > (0 : ℝ), Metric.ball (0 : ℚ) ε ⊆ U i0 := by
    rcases (Metric.isOpen_iff.mp hi0_open) (0 : ℚ) hi0 with ⟨ε, hε_pos, hball_sub⟩
    exact ⟨ε, hε_pos, hball_sub⟩
  obtain ⟨N, hN⟩ : ∃ N : ℕ, 1 / (((N : ℕ) : ℝ) + 1) < ε :=
    exists_nat_one_div_lt hε_pos
  -- All q(n) for n ≥ N land in U(i0) via the ε-ball
  have h_tail_in_U0 (n : ℕ) (hn : N ≤ n) : q n ∈ U i0 := by
    have h_bound : dist (q n) (0 : ℚ) < ε := by
      have hqb := Metric.mem_ball.mp (hq_ball n)
      have h_denom : 1 / (((n : ℕ) : ℝ) + 1) ≤ 1 / (((N : ℕ) : ℝ) + 1) := by
        refine (one_div_le_one_div (by positivity) (by positivity)).mpr ?_
        have h_n_ge_N : (N : ℝ) ≤ (n : ℝ) := by exact mod_cast hn
        linarith
      linarith
    exact hball_sub (Metric.mem_ball.mpr h_bound)
  -- The finitely many q(n) for n < N each have a cover element
  have h_fin_covers_head (n : ℕ) (hn : n < N) : ∃ i : ι, q n ∈ U i := by
    have hqn_mem : q n ∈ ({(0 : ℚ)} ∪ Set.range q) := by simp
    rcases Set.mem_iUnion.mp (hU_cover hqn_mem) with ⟨i, hi⟩
    exact ⟨i, hi⟩
  let idx_fn : ℕ → ι := fun n =>
    if hn : n < N then Classical.choose (h_fin_covers_head n hn) else i0
  let head_indices : Finset ι :=
    Finset.image idx_fn (Finset.range N)
  let t : Finset ι := {i0} ∪ head_indices
  refine ⟨t, ?_⟩
  intro x hxS
  rcases hxS with (rfl | ⟨n, rfl⟩)
  · exact Set.mem_iUnion₂.mpr ⟨i0, Finset.mem_union.mpr (Or.inl (by simp)), hi0⟩
  · by_cases hnN : n < N
    · have hi_mem : idx_fn n ∈ head_indices := by
        dsimp [head_indices]
        exact Finset.mem_image.mpr ⟨n, Finset.mem_range.mpr hnN, rfl⟩
      have hqn_in_U : q n ∈ U (idx_fn n) := by
        dsimp [idx_fn]
        rw [dif_pos hnN]
        exact Classical.choose_spec (h_fin_covers_head n hnN)
      exact Set.mem_iUnion₂.mpr
        ⟨idx_fn n, Finset.mem_union.mpr (Or.inr hi_mem), hqn_in_U⟩
    · have hNle : N ≤ n := by omega
      exact Set.mem_iUnion₂.mpr
        ⟨i0, Finset.mem_union.mpr (Or.inl (by simp)), h_tail_in_U0 n hNle⟩

/--
Prove that no sequence $\{K_n\}_{n=0}^{\infty}$ of compact (closed and bounded) sets of rational numbers has the property that every compact set of rational numbers is contained by at least one $K_n$.
-/
theorem putnam_1968_b6
  : ¬∃ K : ℕ → Set ℚ, (∀ n : ℕ, IsCompact (K n)) ∧ (∀ S : Set ℚ, IsCompact S → ∃ n : ℕ, S ⊆ K n) := by
  intro h
  rcases h with ⟨K, ⟨hK, hcover⟩⟩
  -- For each n, choose q(n) near 0 but outside K(n), using density of Kₙᶜ
  have h_ball_nonempty (n : ℕ) : (Metric.ball (0 : ℚ) (1 / ((n : ℝ) + 1))).Nonempty := by
    rw [Metric.nonempty_ball]
    positivity
  have h_exists_q (n : ℕ) : ∃ q : ℚ, q ∈ Metric.ball (0 : ℚ) (1 / ((n : ℝ) + 1)) ∧ q ∉ K n := by
    have hdense : Dense ((K n)ᶜ) := Rat.dense_compl_compact (hK n)
    have hball_open : IsOpen (Metric.ball (0 : ℚ) (1 / ((n : ℝ) + 1))) := Metric.isOpen_ball
    have h_inter := (dense_iff_inter_open.mp hdense) _ hball_open (h_ball_nonempty n)
    rcases h_inter with ⟨q, hq_mem, hq_not⟩
    exact ⟨q, hq_mem, hq_not⟩
  choose q hq_ball hq_notin using h_exists_q
  -- S = {0} ∪ range(q) is compact (q converges to 0) but avoids every K(n)
  let S : Set ℚ := {0} ∪ Set.range q
  have hS_compact : IsCompact S := isCompact_zero_union_range q hq_ball
  have hS_not_covered (n : ℕ) : ¬ S ⊆ K n := by
    intro h_sub
    exact hq_notin n (h_sub (by simp [S]))
  rcases hcover S hS_compact with ⟨n, hn⟩
  exact hS_not_covered n hn
\end{lstlisting}

\begin{lstlisting}[style=leanlisting,caption={Proof-Refactor output for Putnam 1968 B6.},label={lst:case-pipeline-1968b6}]
import Mathlib

open Finset Polynomial Topology Filter Metric

/-- In the rationals, any nonempty open set contains a point outside any given compact set. -/
lemma Rat.exists_mem_open_not_mem_compact
    {U K : Set ℚ} (hU_open : IsOpen U) (hU_nonempty : U.Nonempty) (hK_compact : IsCompact K) :
    ∃ q ∈ U, q ∉ K := by
  have hdense : Dense (Kᶜ) := Rat.dense_compl_compact hK_compact
  have h_inter := (dense_iff_inter_open.mp hdense) _ hU_open hU_nonempty
  rcases h_inter with ⟨q, hq_U, hq_compl⟩
  exact ⟨q, hq_U, hq_compl⟩

/-- A sequence whose n-th term lies in the ball of radius `1/(n+1)` around a point converges to that point. -/
lemma tendsto_of_mem_ball_one_div_succ
    {α : Type*} [PseudoMetricSpace α] (x : α) (q : ℕ → α)
    (hq : ∀ (n : ℕ), q n ∈ Metric.ball x (1 / ((n : ℝ) + 1))) :
    Filter.Tendsto q Filter.atTop (nhds x) := by
  rw [Metric.tendsto_atTop]
  intro ε hε
  obtain ⟨N, hN⟩ := exists_nat_one_div_lt hε
  refine ⟨N, fun n hn => ?_⟩
  have h1 := Metric.mem_ball.mp (hq n)
  have h2 : (1 : ℝ) / ((↑n : ℝ) + 1) ≤ 1 / ((↑N : ℝ) + 1) := by
    apply one_div_le_one_div_of_le (by positivity)
    have : (↑N : ℝ) ≤ (↑n : ℝ) := by exact_mod_cast hn
    linarith
  linarith

/--
Prove that no sequence $\{K_n\}_{n=0}^{\infty}$ of compact (closed and bounded) sets of rational numbers has the property that every compact set of rational numbers is contained by at least one $K_n$.
-/
theorem putnam_1968_b6
  : ¬∃ K : ℕ → Set ℚ, (∀ n : ℕ, IsCompact (K n)) ∧ (∀ S : Set ℚ, IsCompact S → ∃ n : ℕ, S ⊆ K n) := by
  intro h
  rcases h with ⟨K, ⟨hK, hcover⟩⟩
  have h_ball_nonempty (n : ℕ) : (Metric.ball (0 : ℚ) (1 / ((n : ℝ) + 1))).Nonempty := by
    rw [Metric.nonempty_ball]
    positivity
  have h_exists_q (n : ℕ) : ∃ q : ℚ, q ∈ Metric.ball (0 : ℚ) (1 / ((n : ℝ) + 1)) ∧ q ∉ K n := by
    exact Rat.exists_mem_open_not_mem_compact Metric.isOpen_ball (h_ball_nonempty n) (hK n)
  choose q hq_ball hq_notin using h_exists_q
  have hq_tendsto := tendsto_of_mem_ball_one_div_succ 0 q hq_ball
  let S : Set ℚ := {0} ∪ Set.range q
  have hS_compact : IsCompact S := hq_tendsto.isCompact_insert_range
  have hS_not_covered (n : ℕ) : ¬ S ⊆ K n := by
    intro h_sub
    have hqn_in_Kn : q n ∈ K n := h_sub (by simp [S])
    exact hq_notin n hqn_in_Kn
  have h_contra := hcover S hS_compact
  rcases h_contra with ⟨n, hn⟩
  exact hS_not_covered n hn
\end{lstlisting}

\begin{lstlisting}[style=leanlisting,caption={Original generated proof for Putnam 1971 A2.},label={lst:case-original-1971a2}]
import Mathlib

open Set

abbrev putnam_1971_a2_solution : Set (Polynomial ℝ) := {Polynomial.X}
-- {Polynomial.X}
/--
Determine all polynomials $P(x)$ such that $P(x^2 + 1) = (P(x))^2 + 1$ and $P(0) = 0$.
-/
theorem putnam_1971_a2
    (P : Polynomial ℝ) :
    (P.eval 0 = 0 ∧ (∀ x : ℝ, P.eval (x^2 + 1) = (P.eval x)^2 + 1)) ↔ P ∈ putnam_1971_a2_solution := by
  rw [putnam_1971_a2_solution, Set.mem_singleton_iff]
  constructor
  · intro ⟨h0, heq⟩
    by_contra h_ne
    let a : ℕ → ℝ := λ n => Nat.recOn n 0 (λ _ ak => ak ^ 2 + 1)
    have ha0 : a 0 = 0 := rfl
    have ha_succ : ∀ n, a (n + 1) = (a n) ^ 2 + 1 := λ _ => rfl
    have ha_nonneg : ∀ n, 0 ≤ a n := by
      intro n
      induction' n with n ih
      · rfl
      · rw [ha_succ n]
        nlinarith
    have ha_pos : ∀ n, 0 < a (n + 1) := by
      intro n
      rw [ha_succ n]
      nlinarith [ha_nonneg n]
    have ha_strictMono : ∀ n, a n < a (n + 1) := by
      intro n
      rw [ha_succ n]
      rcases Classical.em (a n = 0) with (hzero | hpos_case)
      · rw [hzero]; norm_num
      · have hpos : 0 < a n := lt_of_le_of_ne (ha_nonneg n) (Ne.symm hpos_case)
        nlinarith
    have ha_inj : Function.Injective a := by
      intro x y h
      rcases Nat.lt_trichotomy x y with (hlt | heq | hlt)
      · have hle : x.succ ≤ y := hlt
        have : a x < a y :=
          Nat.le_induction
            (by simpa [Nat.succ_eq_add_one] using ha_strictMono x)
            (fun k hk ih =>
              lt_trans ih (by simpa [Nat.succ_eq_add_one] using ha_strictMono k))
            y hle
        linarith
      · exact heq
      · have hle : y.succ ≤ x := hlt
        have : a y < a x :=
          Nat.le_induction
            (by simpa [Nat.succ_eq_add_one] using ha_strictMono y)
            (fun k hk ih =>
              lt_trans ih (by simpa [Nat.succ_eq_add_one] using ha_strictMono k))
            x hle
        linarith
    have hP_a : ∀ n, P.eval (a n) = a n := by
      intro n
      induction' n with n ih
      · exact h0
      · rw [ha_succ n, heq (a n), ih]
    have hroot : ∀ n, (P - Polynomial.X).eval (a n) = 0 := by
      intro n
      rw [Polynomial.eval_sub, hP_a n, Polynomial.eval_X, sub_self]
    set Q := P - Polynomial.X with hQ_def
    have hQ_ne : Q ≠ 0 := by
      intro hQzero
      apply h_ne
      rw [hQ_def] at hQzero
      exact sub_eq_zero.mp hQzero
    set d := Q.natDegree with hd_def
    let Z : Finset ℝ := (Finset.range (d + 1)).image a
    have hZ_card : Z.card = d + 1 := by
      rw [Finset.card_image_of_injective _ ha_inj, Finset.card_range]
    have hZ_val_subset : Z.val ⊆ Q.roots := by
      rw [Multiset.subset_iff]
      intro x hx
      rw [Finset.mem_val] at hx
      have hxZ : x ∈ Z := hx
      rw [Polynomial.mem_roots hQ_ne]
      rcases Finset.mem_image.mp hxZ with ⟨n, _, rfl⟩
      dsimp [Q, Polynomial.IsRoot]
      rw [Polynomial.eval_sub, hP_a n, Polynomial.eval_X, sub_self]
    have h_le : Z.card ≤ Q.natDegree :=
      Polynomial.card_le_degree_of_subset_roots hZ_val_subset
    rw [hZ_card, hd_def] at h_le
    omega
  · intro hP
    subst hP
    refine ⟨?_, ?_⟩
    · simp
    · intro x; simp
\end{lstlisting}

\begin{lstlisting}[style=leanlisting,caption={Baseline refactoring for Putnam 1971 A2.},label={lst:case-baseline-1971a2}]
import Mathlib

open Set

abbrev putnam_1971_a2_solution : Set (Polynomial ℝ) := {Polynomial.X}
-- {Polynomial.X}
/--
Determine all polynomials $P(x)$ such that $P(x^2 + 1) = (P(x))^2 + 1$ and $P(0) = 0$.
-/
theorem putnam_1971_a2
    (P : Polynomial ℝ) :
    (P.eval 0 = 0 ∧ (∀ x : ℝ, P.eval (x^2 + 1) = (P.eval x)^2 + 1)) ↔ P ∈ putnam_1971_a2_solution := by
  rw [putnam_1971_a2_solution, Set.mem_singleton_iff]
  constructor
  · intro ⟨h0, heq⟩
    by_contra h_ne
    -- Define the orbit sequence a(0) = 0, a(n+1) = a(n)² + 1
    let a : ℕ → ℝ := λ n => Nat.recOn n 0 (λ _ ak => ak ^ 2 + 1)
    have ha_succ : ∀ n, a (n + 1) = (a n) ^ 2 + 1 := λ _ => rfl
    have ha_nonneg : ∀ n, 0 ≤ a n := by
      intro n
      induction' n with n ih
      · rfl
      · rw [ha_succ n]; nlinarith
    -- The sequence is strictly increasing, hence injective
    have ha_strictMono : ∀ n, a n < a (n + 1) := by
      intro n
      rw [ha_succ n]
      rcases Classical.em (a n = 0) with (hzero | hpos_case)
      · rw [hzero]; norm_num
      · have hpos : 0 < a n := lt_of_le_of_ne (ha_nonneg n) (Ne.symm hpos_case)
        nlinarith
    have ha_inj : Function.Injective a := (strictMono_nat_of_lt_succ ha_strictMono).injective
    -- P fixes every term of the sequence: P(a(n)) = a(n)
    have hP_a : ∀ n, P.eval (a n) = a n := by
      intro n
      induction' n with n ih
      · exact h0
      · rw [ha_succ n, heq (a n), ih]
    -- So Q = P - X vanishes on every a(n)
    have hroot : ∀ n, (P - Polynomial.X).eval (a n) = 0 := by
      intro n
      rw [Polynomial.eval_sub, hP_a n, Polynomial.eval_X, sub_self]
    -- Contradiction: Q has degree d but at least d + 1 distinct roots
    set Q := P - Polynomial.X with hQ_def
    have hQ_ne : Q ≠ 0 := by
      intro hQzero
      apply h_ne
      rw [hQ_def] at hQzero
      exact sub_eq_zero.mp hQzero
    set d := Q.natDegree with hd_def
    let Z : Finset ℝ := (Finset.range (d + 1)).image a
    have hZ_card : Z.card = d + 1 := by
      rw [Finset.card_image_of_injective _ ha_inj, Finset.card_range]
    have hZ_val_subset : Z.val ⊆ Q.roots := by
      rw [Multiset.subset_iff]
      intro x hx
      rw [Finset.mem_val] at hx
      rw [Polynomial.mem_roots hQ_ne]
      rcases Finset.mem_image.mp hx with ⟨n, _, rfl⟩
      exact hroot n
    have h_le : Z.card ≤ Q.natDegree :=
      Polynomial.card_le_degree_of_subset_roots hZ_val_subset
    rw [hZ_card, hd_def] at h_le
    omega
  · intro hP
    subst hP
    refine ⟨?_, ?_⟩
    · simp
    · intro x; simp
\end{lstlisting}

\begin{lstlisting}[style=leanlisting,caption={Proof-Refactor output for Putnam 1971 A2.},label={lst:case-pipeline-1971a2}]
import Mathlib

open Set

abbrev putnam_1971_a2_solution : Set (Polynomial ℝ) := {Polynomial.X}
-- {Polynomial.X}

/-- For a sequence with `a 0 = 0` and `a (n+1) = a n ^ 2 + 1`, every term is nonnegative. -/
private lemma seq_nonneg
    (a : ℕ → ℝ) (ha0 : a 0 = 0)
    (ha_succ : ∀ n, a (n + 1) = a n ^ 2 + 1) (n : ℕ) :
    0 ≤ a n := by
  induction n with
  | zero => simp [ha0]
  | succ n ih => rw [ha_succ]; nlinarith [sq_nonneg (a n)]

/-- A nonnegative sequence satisfying `a (n+1) = a n ^ 2 + 1` is strictly increasing. -/
private lemma seq_strictMono
    (a : ℕ → ℝ) (ha_succ : ∀ n, a (n + 1) = a n ^ 2 + 1)
    (ha_nonneg : ∀ n, 0 ≤ a n) (n : ℕ) :
    a n < a (n + 1) := by
  rw [ha_succ]
  nlinarith [sq_nonneg (a n), ha_nonneg n]

/-- If `P(0) = 0` and `P(x² + 1) = P(x)² + 1`, then `P(aₙ) = aₙ` for the sequence
`a 0 = 0`, `a (n+1) = a n ^ 2 + 1`. -/
private lemma eval_seq_eq_seq {α : Type*} [CommRing α]
    (P : Polynomial α) (a : ℕ → α)
    (h0 : P.eval 0 = 0)
    (heq : ∀ x : α, P.eval (x ^ 2 + 1) = P.eval x ^ 2 + 1)
    (ha0 : a 0 = 0)
    (ha_succ : ∀ n, a (n + 1) = a n ^ 2 + 1) (n : ℕ) :
    P.eval (a n) = a n := by
  induction n with
  | zero => simp [ha0, h0]
  | succ n ih => rw [ha_succ, heq, ih]

/-- The multiset underlying the image of a finite set under `f` is contained in the roots of `p`
whenever `p.eval (f i) = 0` for each `i` in the set. -/
lemma image_val_subset_roots {ι α : Type*} [DecidableEq α] [CommRing α] [IsDomain α]
    {p : Polynomial α} (hp : p ≠ 0) {f : ι → α} {s : Finset ι}
    (h_root : ∀ i ∈ s, p.eval (f i) = 0) :
    (s.image f).val ⊆ p.roots := by
  intro x hx
  rw [Polynomial.mem_roots hp]
  obtain ⟨i, hi, rfl⟩ := Finset.mem_image.mp (Finset.mem_val.mp hx)
  exact h_root i hi

/--
Determine all polynomials $P(x)$ such that $P(x^2 + 1) = (P(x))^2 + 1$ and $P(0) = 0$.
-/
theorem putnam_1971_a2
    (P : Polynomial ℝ) :
    (P.eval 0 = 0 ∧ (∀ x : ℝ, P.eval (x^2 + 1) = (P.eval x)^2 + 1)) ↔ P ∈ putnam_1971_a2_solution := by
  rw [putnam_1971_a2_solution, Set.mem_singleton_iff]
  constructor
  · intro ⟨h0, heq⟩
    by_contra h_ne
    let a : ℕ → ℝ := λ n => Nat.recOn n 0 (λ _ ak => ak ^ 2 + 1)
    have ha_succ : ∀ n, a (n + 1) = (a n) ^ 2 + 1 := λ _ => rfl
    have ha_nonneg : ∀ n, 0 ≤ a n := seq_nonneg a rfl ha_succ
    have ha_strictMono : ∀ n, a n < a (n + 1) := seq_strictMono a ha_succ ha_nonneg
    have ha_inj : Function.Injective a :=
      (strictMono_nat_of_lt_succ ha_strictMono).injective
    have hP_a : ∀ n, P.eval (a n) = a n := eval_seq_eq_seq P a h0 heq rfl ha_succ
    set Q := P - Polynomial.X with hQ_def
    have hQ_ne : Q ≠ 0 := by
      intro hQzero
      apply h_ne
      rw [hQ_def] at hQzero
      exact sub_eq_zero.mp hQzero
    set d := Q.natDegree with hd_def
    let Z : Finset ℝ := (Finset.range (d + 1)).image a
    have hZ_card : Z.card = d + 1 := by
      rw [Finset.card_image_of_injective _ ha_inj, Finset.card_range]
    have hZ_val_subset : Z.val ⊆ Q.roots := by
      apply image_val_subset_roots hQ_ne
      intro i _
      show Polynomial.eval (a i) (P - Polynomial.X) = 0
      rw [Polynomial.eval_sub, Polynomial.eval_X, hP_a i, sub_self]
    have h_le : Z.card ≤ Q.natDegree :=
      Polynomial.card_le_degree_of_subset_roots hZ_val_subset
    rw [hZ_card, hd_def] at h_le
    omega
  · intro hP
    subst hP
    refine ⟨?_, ?_⟩
    · simp
    · intro x; simp
\end{lstlisting}

\clearpage
\subsection{Evaluation Rubric}
\label{app:evaluation-rubric}

The listing below records the rubric used by the LLM-as-a-judge evaluator.

\begin{lstlisting}[style=promptlisting,caption={Evaluation rubric.},label={lst:evaluation-rubric}]
===== DETAILED EVALUATION RUBRIC =====

The proof files are assumed to have already passed Lean verification. Do not score correctness. Score only refactor quality: structure, helper quality, tactic transparency, reuse, and human readability.

All scores are from 1.0 to 5.0, where higher is better. Half-points such as 3.5 or 4.5 are allowed. Be consistent across methods. When comparing benchmark_refactor and proof_refactor_pipeline, calibrate their scores relative to each other.

Do not reward theorem count or lemma count by itself. Do not reward length reduction by itself. Penalize fake modularity: wrappers, dead helpers, huge helpers, or tactic-golf helpers. Prefer meaningful, named, locally checkable mathematical lemmas. A single-use helper is acceptable if it exposes real structure. Broad tactics are acceptable only when they close local routine goals, not when they hide the main proof idea.

## structure

Required metrics:

- main_theorem_slimness
- complexity_distribution
- dependency_clarity

Meaning:

- main_theorem_slimness: whether the main theorem has become a readable proof outline rather than a monolithic proof body.
- complexity_distribution: whether proof complexity is distributed into meaningful helper lemmas rather than hidden in one huge helper.
- dependency_clarity: whether the dependency relation among helpers and the main theorem is clear and shallow.

Scoring guide:

- 5: main proof is slim, helper structure is balanced, dependencies are clear.
- 4: good structure with minor imbalance.
- 3: some structural improvement, but still proof-script-heavy or uneven.
- 2: main proof is shorter mostly by hiding complexity.
- 1: structure is worse or obviously metric-gamed.

Inspection checklist:

- Inspect top-level helper lemmas.
- Inspect private helper lemmas.
- Inspect the main theorem body.
- Inspect local `have` blocks.
- Judge whether the main theorem reads like a proof outline.
- Judge whether helper lemmas carry meaningful proof structure.

## signature_quality

Required metrics:

- statement_naturalness
- binder_economy
- generality

Meaning:

- statement_naturalness: whether helper statements look like meaningful mathematical lemmas.
- binder_economy: whether helper signatures avoid excessive local hypotheses and unnecessary parameters.
- generality: whether helper lemmas are generalized beyond a local proof fragment when appropriate.

Note that: If two helpers' signatures have the same conclusion, but one has more general assumptions, the more general one is better and add extra assumptions should be penalized. 





Scoring guide:

- 5: helper statements are natural, concise, and reusable.
- 4: mostly natural, with some theorem-specific statements.
- 3: usable but local or proof-script-like.
- 2: many over-specialized helpers.
- 1: helpers are mostly meaningless wrappers or fragments.

Inspection checklist:

- Judge whether each helper name expresses mathematical content.
- Judge whether each helper statement is natural.
- Judge whether the statement is too local to the original proof.
- Judge whether it has too many binders or hypotheses.
- Judge whether it is more general than the immediate scaffold or merely copied from it.

## tactic_quality

Required metrics:

- tactic_transparency
- explicit_lemma_use
- broad_tactic_control

Meaning:

- tactic_transparency: whether proof bodies use understandable local reasoning rather than opaque closures.
- explicit_lemma_use: whether the refactored proof actually calls named helpers in meaningful places.
- broad_tactic_control: whether broad tactics such as `simp_all`, `aesop`, `grind`, `omega`, or large `nlinarith` calls are controlled and not hiding major structure.

Do not penalize all automation. Local uses of `ring`, `linarith`, `norm_num`, `field_simp`, `positivity`, or explicit `simp [lemmas]` may be good style. A typical good use case is 
to use this tactic to prove some straightforward algebraic manipulation or inequality
or logical tautology but tedious.


Benign or usually acceptable tactics:

```lean
ring
norm_num
field_simp
positivity
linarith [explicit facts]
simp [specific_lemma]
rw [specific_lemma]
```

Potentially broad or opaque tactics:

```lean
simp_all
aesop
grind
omega
large nlinarith without explicit structure
```

Scoring guide:

- 5: explicit lemma use dominates; broad tactics are rare and local.
- 4: some broad tactics, but they do not hide major proof structure.
- 3: moderate opaque closure.
- 2: many helpers are closed by broad tactics.
- 1: proof is mostly tactic golf.

Inspection checklist:

- Judge whether broad tactics are used only for local routine goals.
- Judge whether named helpers are actually called in meaningful places.
- Judge whether the proof hides the main idea behind automation.

## reuse

Required metrics:

- helper_usefulness
- reuse_potential
- no_dead_helpers

Meaning:

- helper_usefulness: whether helpers play real structural roles in the proof.
- reuse_potential: whether some helpers could plausibly be reused outside the exact local proof position.
- no_dead_helpers: whether there are no unused, orphan, or purely decorative helpers.

Scoring guide:

- 5: helpers are clearly useful; several have reuse potential; no dead helpers.
- 4: all helpers are justified; some reuse potential.
- 3: mostly single-use helpers, but structurally useful.
- 2: many single-use helpers with weak semantic value.
- 1: unused or fake helpers are common.

Inspection checklist:

- Check whether helpers are actually used.
- Check whether single-use helpers are structurally meaningful.
- Check whether helpers are reused by multiple parts of the proof.
- Check whether any helpers are unused or merely decorative.
- Check whether any helpers are wrappers around another lemma without real abstraction.

## human_review

Required metrics:

- proof_readability
- mathlib_style
- maintainability

Meaning:

- proof_readability: whether a human can read the proof as a clear mathematical argument.
- mathlib_style: whether names, locality, statement shape, binder order, and tactic usage are close to Mathlib style.
- maintainability: whether future changes would likely be localized and easy to repair.

Mathlib style is general as much as possible, if you generalize type or typeclass assumptions, e.g form `f : ℝ → ℝ` to `f : α → β` with some typeclass assumptions, it is better. And if you find the proof use more mathlib lemmas for those new proof, it is better. 

Scoring guide:

- 5: very readable, Mathlib-like, easy to maintain.
- 4: good proof-engineering style.
- 3: acceptable but still somewhat ad hoc.
- 2: hard to maintain.
- 1: barely readable despite compiling.

Inspection checklist:

- Judge whether a human reader can recover the mathematical proof outline.
- Judge whether names, binder order, locality, and statement shape are close to Mathlib style.
- Judge whether future proof repairs would likely be localized.

## overall

Compute overall.score as the simple average of all 15 non-overall metric scores for the method, rounded to 2 decimals.

## Output requirement

Return only the required JSON object. Each method must include a `reason` string. The `reason` should be concise but substantive, usually 3-6 sentences, explaining why the method received its scores. Do not put Markdown outside the JSON.
\end{lstlisting}

% The NeurIPS checklist is intentionally omitted from the arXiv source.

\end{document}